\documentclass[letterpaper, 10 pt, conference]{ieeeconf}  

\IEEEoverridecommandlockouts                              
\newcommand{\ie}{\textit{i}.\textit{e}.}
\usepackage{mathrsfs}
\usepackage{amsfonts}
\usepackage{amsmath,graphicx}
\usepackage{epstopdf}
\usepackage{color}
\usepackage{wrapfig}
\usepackage{lipsum}
\usepackage{float}
\usepackage{colortbl,booktabs}
\usepackage{mathtools}
\usepackage{cite}
\usepackage{subfigure}
\usepackage{multirow}
\title{\LARGE \bf
End-to-end Learning for Inter-Vehicle Distance and Relative Velocity Estimation in ADAS with a Monocular Camera
}
\author{Zhenbo Song$^{1}$  Jianfeng Lu$^{1}$  Tong Zhang$^{2}$  Hongdong Li$^{3}$
\thanks{$^{1}$ Zhenbo Song and Jianfeng Lu are with the School of Computer Science and Engineering,
        Nanjing University of Science and Technology, 210094 Nanjing, China.
        {\tt\small [songzb,lujf]@njust.edu.cn}}%
\thanks{$^{2}$Tong Zhang is with Research School of Engineering, Australian National University and Motovis Australia Pty Ltd. 
          {\tt\small tong.zhang@anu.edu.au }} %
\thanks{$^{3}$Hongdong Li is with Research School of Engineering,
        Australian National University, 0200 Canberra, Australia.
        {\tt\small  hongdong.li@anu.edu.au}}%
}

\begin{document}

\maketitle
\thispagestyle{empty}
\pagestyle{empty}

\begin{abstract}
Inter-vehicle distance and relative velocity estimations are two basic functions for any ADAS (Advanced driver-assistance systems). In this paper, we propose a monocular camera based inter-vehicle distance and relative velocity estimation method based on end-to-end training of a deep neural network. The key novelty of our method is the integration of multiple visual clues provided by any two time-consecutive monocular frames, which include deep feature clue, scene geometry clue, as well as temporal optical flow clue.  We also propose a vehicle-centric sampling mechanism to alleviate the effect of perspective distortion in the motion field (\ie  optical flow). We implement the method by a light-weight deep neural network. Extensive experiments are conducted which confirm the superior performance of our method over other state-of-the-art methods,in terms of estimation accuracy, computational speed, and memory footprint. 
\end{abstract}

\IEEEpeerreviewmaketitle

\section{Introduction}
Advanced driver assistance systems (or ADAS) focus on the active safety technologies that alert the driver to potential problems or to avoid collisions automatically.  Inter-vehicle distance estimation and relative velocity estimation are two essential capabilities required by any modern ADAS.

Applying range sensors (e.g. LiDAR or radar) is one of the most representative solutions in ADAS applications. These sensors provide direct measurements of other vehicles' distance and velocity, however, they are susceptible to adverse environment factors such as rain, snow or fog~\cite{hasirlioglu2017effects}. 
By contrast, camera sensors offer richer texture and structure information of the scene even in adverse conditions, which is considered as a cost-effective and powerful alternative to range sensors.
Meanwhile, deep learning has recently made a great success in many visual applications, such object detection~\cite{ren2015faster, redmon2016you}, optical flow estimation~\cite{dosovitskiy2015flownet, sun2018pwc} and depth prediction~\cite{FuCVPR18-DORN, bian2019depth}.
Considering the economic efficiency and environmental adaptation, we therefore explore a new deep learning method to estimate the distance and velocity using a monocular camera.

Previous works address the distance regression problem from two different perspectives: monocular depth estimation and 3D object detection. It has been debated whether monocular depth estimation is capable of producing reliable distance estimation. The studies~\cite{FuCVPR18-DORN, bian2019depth, Yin2019enforcing} pointed out that predicted depth maps are lack of an unknown scale factor in some datasets, which is a main obstacle to distance regression. Subsequently, ~\cite{wang2018monocular} proposes to tackle the scale ambiguity by exploiting geometric constraints between road surface and the camera height, although it introduces extra processing for road surface estimation.  On the other hand, 3D object detection\cite{mousavian20173d, brazil2019m3drpn, wang2019pseudo, qin2019tlnet} in recent years focus on recovering both 6-DoF pose and dimensions of an object from an image. Existing 2D detection algorithms have demonstrated their capability to deal with large variations in viewpoint and clutter. Hence, 3D detection leverages the power of 2D detection to guide and improve the performance.It has been noted from 3D detection works that deep features of vehicles contain the characteristics of the vehicle itself, including orientation, dimension, observation angle and even key points of vehicle's 3D model.

Inspired by 3D bounding box estimation~\cite{mousavian20173d}, we propose our vehicle distance regression model based on 2D detection by assuming that the 2D bounding boxes tightly surround corresponding 3D box projections.
To simply adapt our method to practical applications, we regress the closest distance to the vehicle rather than the whole 3D bounding box.
Our approach takes deep features and homogeneous width and height of detected 2D bounding boxes as input, where deep features offer real scale characteristics of the vehicle and homogeneous width and height constrain the proportional relation between the vehicle's distance and its characteristics.

Bringing the definition of scene flow~\cite{menze2015object}, the relative velocity is considered as the 3D motion of other vehicle centers relative to the measuring camera during a time interval. Recent stereo video sequence based object scene flow estimation approaches~\cite{ma2019deep, vogel20153d} achieve impressive results, though, employing stereo video suffering from very high computational cost. Therefore, monocular camera along with different clues have been proposed to estimate the vehicle velocity in real time scenarios. For example, benefited from the stationary camera setting and the road surface constraint, surveillance cameras are used to analyse traffic flow and vehicle velocity~\cite{tran2018traffic}. However, the estimation becomes fragile in dynamic scene where it is hard to get the real depth of vehicles and map the temporal clues in images to the velocity in reality. There are few works working on monocular velocity estimation, except for~\cite{kampelmuhler2018camera}, which regresses the relative velocity of other vehicles via multiple complex clues including vehicle tracking, dense depth and optical flow information. Nevertheless, we are arguing that using those redundant information in turn affect the speed of estimation, and better velocity estimation can be achieved with less information. Along with our distance regression model and optical flow estimation network, we design our inter-vehicle relative velocity system in a principled way.

In our dynamic application scenarios, it is not an ideal choice to predict the optical flow in the whole image due to the unbalance motion distribution between the static background and moving vehicles.
When the real flow of a moving vehicle is only a small fraction of pixel, the predicted flow from full image flow networks ends up in zero.
Furthermore, on account of perspective projection, vehicles in distance with high velocity may have small optical flow; on the other hand, close vehicles with low velocity may have large flow on image.
This circumstance enlarges the disadvantage of the full image flow networks. To make our network focus on the prediction of flow on each vehicle correctly, we employ vehicle bounding boxes to crop vehicle-centric template patches.

In a nutshell, our contributions in this paper are two-fold:
\begin{enumerate}
\item[-] A light-weighted inter-vehicle distance and relative velocity regression network is proposed. Our method utilizes the 2D bounding box detection, which allow us to combine the size of the bounding box with deep features as well as the contracted temporal clue.
\item[-] We develop a vehicle-centric sampling method to reduce the impact of motion and perspective on optical flow clue estimation
\end{enumerate}
Empirically,  evaluations on several standard popular datasets  such  as  Tusimple  velocity and KITTI demonstrate that our method achieves the state-of-the-art performance.

\section{Related Work}
{\bf Inter-vehicle Distance estimation }
Two types of deep learning methods exist for solving the inter-vehicle distance estimation task: monocular depth estimation and 3D object detection.

Xu \emph{et~al.}~\cite{xu2017multi} introduce a U-net-structured network to predict dense depth by supervised learning, which fuses information from multi-scale layers and obtains the integration from continuous Conditional Random Fields.
Instead of regressing the depth directly,~DORN \cite{FuCVPR18-DORN} considers it as an ordinal regression problem by discretizing depth.
Additional geometric constraints between surface normal and depth are proposed in~\cite{Yin2019enforcing}, where a new loss term is used in training process.
In~\cite{bian2019depth}, both depth map and ego-motion are learnt from image sequences in an unsupervised manner, yet the depth is up to an unknown scale. 

On the other hand, given the geometric constraints provided by 2d object bounding box, Mousavian~\emph{et~al.}~\cite{mousavian20173d} propose to obtain 3d bounding box estimation from a single image. M3D-RPN~\cite{brazil2019m3drpn} breaks the barrier between 2D and 3D detection by constructing a 3D region proposal network which allows 3D boxes to utilize convolutional features generated in the 2D image-space. In addition to, triangulation processing in multiple image frames is successfully applied on 3d bounding boxes~\cite{qin2019tlnet}. Similarly, it~\cite{wang2019pseudo} converts the depth map predicted from single image to pseudo-LiDAR point clouds to mimick the LiDAR signal, then feeds LiDAR-based 3D object detectors to form a full pipeline of single image 3D detection.


{\bf Relative Velocity estimation}
Along with the survey on the latest progress of camera-based vehicle velocity estimation, two related researching directions are also reviewed: 3D scene flow and 2D optical flow.

Scene flow~\cite{menze2015object} represents the 3D motion field of each point in a image.
Conventional methods~\cite{vogel2013piecewise,vogel20153d} decompose the scene into piece-wise rigid motion planes and solve the discrete continuous optimization problem. At the same time, learning based method~\cite{ma2019deep} exploits to generalize from multiple visual cues in stereo sequences, such as optical flow, semantic segmentation and stereo disparity.

For optical flow estimation, starting with Flownet~\cite{dosovitskiy2015flownet}, a number of end-to-end deep regression models have been proposed~\cite{dosovitskiy2015flownet, sun2018pwc, meister2018unflow}.
To handle large displacement and refine the flow, Flownet2~\cite{ilg2017flownet} stacks multiple networks and utilizes the warping operation, yet resulting in a very large network.
PWC-Net~\cite{sun2018pwc} extends the idea of traditional spatial pyramid and incorporate the feature pyramid processing, warping and cost volume concepts into the network, which leads to a lighter and faster neural network.

For camera-based vehicle relative velocity estimation in~\cite{tran2018traffic}, a stationary camera is approximately calibrated based on the landmarks on the road like broken white painted lines.
The velocity is then measured according to the calibration parameters after tracking the specific vehicle.
Another method~\cite{kampelmuhler2018camera} for autonomous driving regresses the relative velocity through the trajectory features extracted from Monodepth~\cite{godard2017unsupervised}, MedianFlow tracker~\cite{bakerlucas, kalal2010forward} and Flownet~\cite{dosovitskiy2015flownet}

\begin{figure}[h]
\centering
\includegraphics[width=0.5\textwidth]{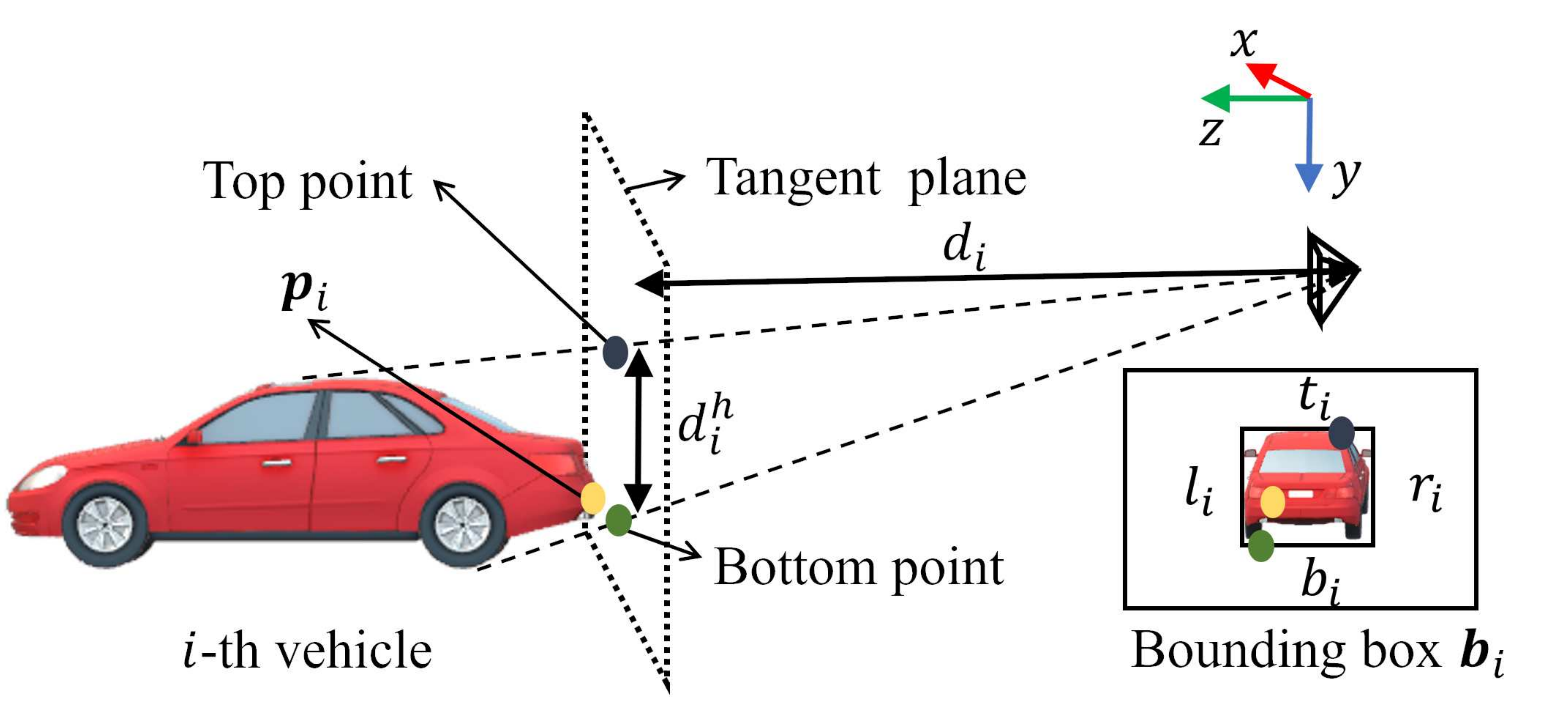}
\caption{Illustration of the perspective projection. The tangent plane is parallel to image plane. ${{\bf p}_i}$ is the closest point on vehicle. ${d_i}$ denotes the inter-vehicle distance. The top point and bottom point are on the tangent plane and correspond to ${t_i}$ and ${b_i}$ coordinates of the bounding box ${{\bf b}_i}$ on image.}
\label{fig.illustration}
\end{figure}

\begin{figure*}[htbp]
\centering
\includegraphics[width=1\textwidth]{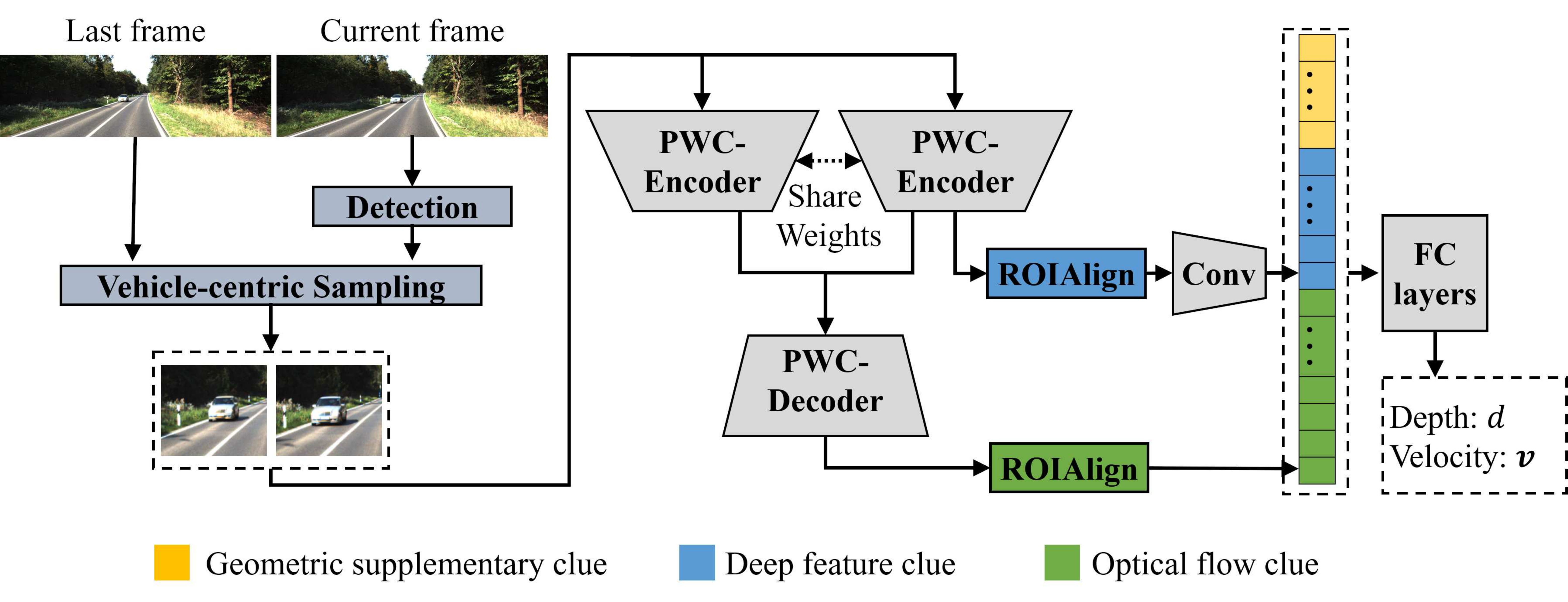}
\caption{Architecture overview of our proposed method. The original image is sampled by \textbf{Vehicle-centric Sampling} according to the \textbf{Detection} results. Then the deep feature clue and optical flow clue are obtained by \textbf{ROIAlign} and aggregation after \textbf{PWC-Encoder} and \textbf{PWC-Decoder}. Adding with geometric supplementary clue, distance and relative motion are regressed by \textbf{FC layers} (fully connection layers). The relative velocity is calculated from relative motion divided by a time interval ${\Delta t}$.}
\label{fig.method}
\end{figure*}

\section{Problem statement}
Given two RGB images ${\mathcal{I}',\mathcal{I}}$ between ${\Delta t}$ time interval from a calibrated monocular camera, our goal is to estimate the distances and velocities of vehicles in the current frame ${\mathcal{I}}$ relative to the camera coordinate system.
Vehicles (car, truck, van etc.) in ${\mathcal{I}}$ are detected by object detector (e.g. Faster-RCNN \cite{ren2015faster}, YOLO \cite{redmon2016you}) as bounding boxes ${\{{\bf b}_i | i=1,...,n\}}$, where each bounding box consists of left, top, right, bottom image coordinates ${{\bf b}_i = (l_i, t_i, r_i, b_i) \in \mathbb{R}^4}$.
We first denote the inter-vehicle distances and relative velocities as ${\{d_i, {\bf v}_i | i=1,...,n\}}$ with ${d_i \in \mathbb{R}^+}$ and ${{\bf v}_i \in \mathbb{R}^3}$ respectively.
As shown in Fig. \ref{fig.illustration}, the camera coordinate system is defined as ${z}$ along the optical axis, ${x}$ parallel to image surface towards right and ${y}$ going down.
 ${d_i}$ represents the inter-vehicle distance from camera optical center to the closest tangent plane on the ${i}$-th vehicle surface, which is orthogonal to the camera optical axis.
We take the intersection point ${{\bf p}_i \in \mathbb{R}^3}$ as the closest point on the vehicle and assume that the corresponding point ${{\bf p}'_i \in \mathbb{R}^3}$ in the last frame ${\mathcal{I}'}$ keeps same on the vehicle surface for ${\triangle t}$ time.
Then the relative velocity is defined as the closest point rather than the vehicle center as ${{\bf v}_i = ({\bf p}_i-{\bf p}'_i)/\Delta t }$. In next section, we introduce how ${d_i}$ and ${{\bf v}_i}$ are regressed using different clues learnt from deep neural network.

\section{Method}
As shown in Fig. \ref{fig.method}, the architecture of our deep neural network is based on PWC-Net \cite{sun2018pwc} since it is a computational efficient optical flow estimation network.
Different clues are extracted from different layers of the network to estimate vehicle distance and velocity.
First, we introduce the distance regression model using deep feature clue and spatial geometric clue of vehicles.
After that, the velocity estimation method is proposed with additional geometric clues and temporal optical flow clue.
At last, we detail the vehicle centric network pipeline to reduce the perspective and motion influence.

\subsection{Distance Regression}
As depicted in Fig.\ref{fig.illustration}, the vehicle surface is projected into a bounding box ${{\bf b}_i}$, where the top point and bottom point of the projection on the tangent plane correspond to ${t_i, b_i}$ respectively.
Let ${d^h_i}$ denote the distance between the top point and bottom point along ${y}$-axis, and according to the perspective projection rule of pinhole camera model, the distance ${d_i}$ can be computed as: 
 \begin{equation}\label{equ.project}
{d_i} = {\frac{f_y * d^h_i}{b_i - t_i}} = {\frac{f_x * d^w_i}{r_i - l_i}},
\end{equation}
where ${f_x, f_y}$ are the focal length and ${d^w_i}$ is the distance between projected points on the tangent plane corresponding to bounding box ${l_i, r_i}$ along ${x}$-axis.
Equation~\eqref{equ.project} consist of two parts. The first part ${{f_y}/{(b_i - t_i)}, {f_x}/{(r_i - l_i)}}$, which constrains the scale between distance and physical size of objects, is a geometric clue directly given by camera intrinsic parameters and 2D bounding box.
The other part ${d^w_i, d^h_i}$ is the physical size of the vehicle's projection on the tangent plane and it only depends on characteristics of the vehicle itself.
Since these vehicle characteristics can be learnt through a large number of training samples, we extract a feature vector ${{\bf f}_i}$ from a deep neural network to represent them implicitly.  
Consequently, to get the distance ${d_i}$, we take the geometric clue as well as the deep feature vector into consideration.

In our network, after cropping and resizing the vehicle region from the last layer of the PWC-Net encoder using ROIAlign \cite{he2017mask}, the feature vector ${{\bf f}_i}$ is aggregated by two convolution layers.
By considering the distance estimation as a regression problem, ${d_i}$ is obtained through several fully connected layers, which takes deep feature clue ${{\bf f}_i}$ and geometric clues  ${{f_y}/{(b_i - t_i)}}$, ${{f_x}/{(r_i - l_i)}}$ as input.
Accordingly, we formulate the distance regression function ${F_{dist}}$ as follows:
\begin{equation}\label{equ.distance}
{d_i} = {F_{dist}(\frac{f_x}{r_i - l_i}, \frac{f_y}{b_i - t_i}, {\bf f}_i; {\bf w}_{dist})},
\end{equation}
Here, ${{\bf w}_{dist}}$ is model parameters of the ${F_{dist}}$ function.

\subsection{Velocity Estimation}
Supposing we already have the distance ${d_i}$ of the ${i}$-th vehicle, the current velocity ${{\bf v}_i}$ is calculated as:
\begin{equation}\label{equ.direct}
{{\bf v}_i} = \frac{{\bf p}_i-{\bf p}'_i}{\Delta t} 
=\frac{1}{\Delta t} \left[ \begin{array}{c}
d_i \frac{u_i-c_x}{f_x} - d'_i \frac{u'_i-c_x}{f_x} \\
d_i \frac{v_i-c_y}{f_y} - d'_i \frac{v'_i-c_y}{f_y} \\
d_i - d'_i
\end{array}\right]
\end{equation}
where ${(c_x, c_y)}$ is the principle point of camera, ${d'_i}$ is the vehicle distance in the last frame ${\mathcal{I}'}$, ${(u_i, v_i)}$, ${(u'_i, v'_i)}$ are the projected image coordinates of ${{\bf p}_i}$ and ${{\bf p}'_i}$ respectively.
Similar to Equation. \eqref{equ.distance}, ${d'_i}$ is a function of the deep feature ${{\bf f}'_i}$ and the homogeneous size of the tracking bounding box ${{\bf b}'_i}$ in the last frame.
Since optical flow contains information of the object motion, both ${{\bf f}'_i}$ and homogeneous size of ${{\bf b}'_i}$ can be induced from the optical flow with the current size and deep feature.
Besides, ${(u'_i, v'_i)}$ can also be obtained by optical flow with known ${(u_i, v_i)}$, which is thought to be inside the bounding box and its location is determined by vehicle characteristics.
Using the current bounding box ${{\bf b}_i}$ and deep feature ${{\bf f}_i}$ together, ${(u_i, v_i)}$ can be represented implicitly.
As a result, on the basis of distance regression model \eqref{equ.distance}, the velocity is estimated by providing two additional clue: the current bounding box and optical flow clue.

Instead of inputting the bounding box directly, we utilize its homogeneous coordinates as another geometric clue combined with intrinsic parameters of the camera.
In order to get optical flow clue, ROIAlign is applied in every flow pyramid layer from PWC-Net decoder to obtain multi-scale flow patches of the vehicle.
Then these flow patches are concatenated and reconstructed into a flow clue vector ${{\bf m}_i}$.
Altogether, all extracted clues are listed bellow:
\begin{itemize}
  \item [1)]
  deep feature vector ${{\bf f}_i}$
  \item [2)]
  optical flow vector ${{\bf m}_i}$
  \item [3)]
  supplementary geometric vector

  ${{\bf g}_i = [ \frac{f_x}{r_i - l_i}, \frac{f_y}{b_i - t_i}, \frac{l_i-c_x}{f_x}, \frac{t_i-c_y}{f_y}, \frac{r_i-c_x}{f_x}, \frac{b_i-c_y}{f_y} ]}$
\end{itemize}
Finally, the velocity is obtained from these clues by a regression model ${F_{velo}}$ with model parameters ${{\bf w}_{velo}}$ as:
\begin{equation}\label{equ.velocity}
{{\bf v}_i} = {\frac{1}{\Delta t} F_{velo}({\bf g}_i, {\bf f}_i, {\bf m}_i; {\bf w}_{velo})},
\end{equation}

\subsection{Vehicle-centric Network}
As mentioned above, we propose the vehicle-centric sampling strategy to handle the unbalance motion distribution and perspective influence.
Taking the bounding box ${{\bf b}_i = (l_i, t_i, r_i, b_i)}$ on the current frame as an example, the cropping location is defined as follows:
\begin{equation}\label{equ.crop}
{{\bf b}^{crop}_i} = {(l_i - \Delta w_i, t_i - \Delta h_i, r_i + \Delta w_i, b_i + \Delta h_i)},
\end{equation}
where ${\Delta w_i = \frac{r_i-l_i}{2} + \delta}$, ${\Delta h_i = \frac{b_i-t_i}{2} + \delta}$ and ${\delta}$ is a small expanding factor.
The cropped patch is resized to a fixed size ${w_r \times h_r }$ afterwards.
In the same way the corresponding template patch is generated from the last frame at the same location.
Then optical flow and deep feature are predicted from the pair of cropped and resized template patches using PWC-Net.

In our complete network, distance regression model ${F_{dist}}$ and velocity estimation model ${F_{velo}}$ are fused together by a fusion model:
\begin{equation}\label{equ.fusion}
{{d_i}, {\Delta t}\cdot{\bf v}_i} = {F_{fu}({\bf g}_i, {\bf f}_i, {\bf m}_i; {\bf w}_{fu})},
\end{equation}
where ${F_{fu}}$ consists of several fully connected layers with the activation function ReLU.
The deep feature clue ${{\bf f}_i}$ and a optical flow vector are predicted from vehicle-centric patches, while the geometric clue ${{\bf g}_i}$ is calculated using the original bounding box coordinates.
Then, the optical flow vector is rescaled referring to the original image size to get the final flow clue ${{\bf m}_i}$.
Our vehicle-centric network utilizes MSE loss for both distance and velocity regression with ground truth supervision.
The total loss is as follows:
\begin{equation}\label{equ.loss}
{L} = {\alpha L_{dist} + \beta L_{velo}},
\end{equation}
where the coefficients ${\alpha}$ and ${\beta}$ are hyperparameters that are manually set for distance and velocity terms in the loss function.

\begin{table*}[htbp]
  \centering
  \caption{Velocity results on Tusimple benchmark. \emph{Rank1} \cite{kampelmuhler2018camera} was the winning approach. \emph{\textbf{ours} org} is our model using the original images; \emph{\textbf{ours} full} is our vehicle-centric model.}
  \label{tb.leaderboard}
  \begin{tabular}{c|c|c|c|c|c}
    \hline
    \multirow{2}*{Methods} & Position & \multicolumn{4}{c}{velocity}\\
                             \cline{2-6}
		                  ~ & MSE & MSE(near) & MSE(medium)  & MSE(far) & MSE(average)\\
    \hline
    \emph{Rank1} \cite{kampelmuhler2018camera} & - &    0.18 &     0.66 &    3.07  & 1.30\\
    \hline
    \emph{Rank2} & - &     0.25 &     0.75 &     3.50 & 1.50 \\
    \hline
    \emph{Rank3} & - &   0.55 &    2.21 &    5.94 & 2.90 \\
    \hline
    \emph{\textbf{ours} org} &  \textbf{9.72} &    0.23 &     0.99 &     3.27 & 1.50 \\
    \hline
    \emph{\textbf{ours} full} &  10.23 &    \textbf{0.15} &    \textbf{ 0.34} &     \textbf{2.09} & \textbf{0.86} \\
    \hline
    \end{tabular}
\end{table*}

\section{Experiment}
We conduct experiments on Tusimple velocity dataset \cite{kampelmuhler2018camera} and  KITTI raw dataset \cite{geiger2012we} for evaluating the performance of our proposed method.
In Tusimple dataset, video sequences are captured at 20fps and each clip is 40 frames long.
Designated vehicles in distance ranging from 5 meters to up to 90 meters are annotated as bounding boxes on the last frame, as well as their ground truth velocity and position generated by range sensors.
KITTI raw dataset has a tracklet of vehicles in each sequence, while providing a set of 3D point clouds in each frame.
To construct our training and testing data, the distance is obtained through 3D point clouds and the velocity is calculated by using the tracking information of tracklets.

{\bf Implementation Details  }
Our network is based on PWC-Net pretrained from FlyingChairs \cite{dosovitskiy2015flownet} and ${7 \times 7}$ ROIAlign is used to unify the size of feature map and flow map.
Two convolution layers for aggregating the deep feature are in size of ${3 \times 3}$ and ${7 \times 7}$ respectively.
In the vehicle-centric sampling step, all patches are resized to ${384 \times 448}$ which matches the size of images PWC-Net pretrained on.
After concatenating the geometric vector, deep feature vector and flow vector, 4 fully connected layers with ReLU activation function are employed to compute the distance and velocity.
The whole network is implemented in PyTorch and trained end-to-end by optimizing the loss function \eqref{equ.loss} with ADAM \cite{kingma2014adam}.
The parameters ${\alpha, \beta}$ in loss function are empirically set as 0.1 and 1.
There are totally 120 epoches preformed with the learning rate of ${1 \times 10^{-4}}$, decay rate of 0.2 and decay step of every 30 epoches.
The model is trained on a PC with an Intel i7, 12G RAM, and a single NVIDIA GTX 1080ti.

{\bf Evaluation Metrics  }
We employ the scene depth metrics from previous works \cite{liu2015learning, eigen2014depth} for the vehicle distance evaluation.
For the velocity evaluation, we follow the TuSimple Velocity Estimation Challenge rules. They first splits individual vehicles into three groups according to their relative distances: near range (${d < 20m}$); medium range (${20m < d < 45m}$); far range (${d>45m}$).
Then the mean square error (MSE) of velocity is calculated in each range group and the overall average MSE of three groups is the last metric for ranking.

\begin{figure}[htbp]
\centering
\includegraphics[width=0.5\textwidth]{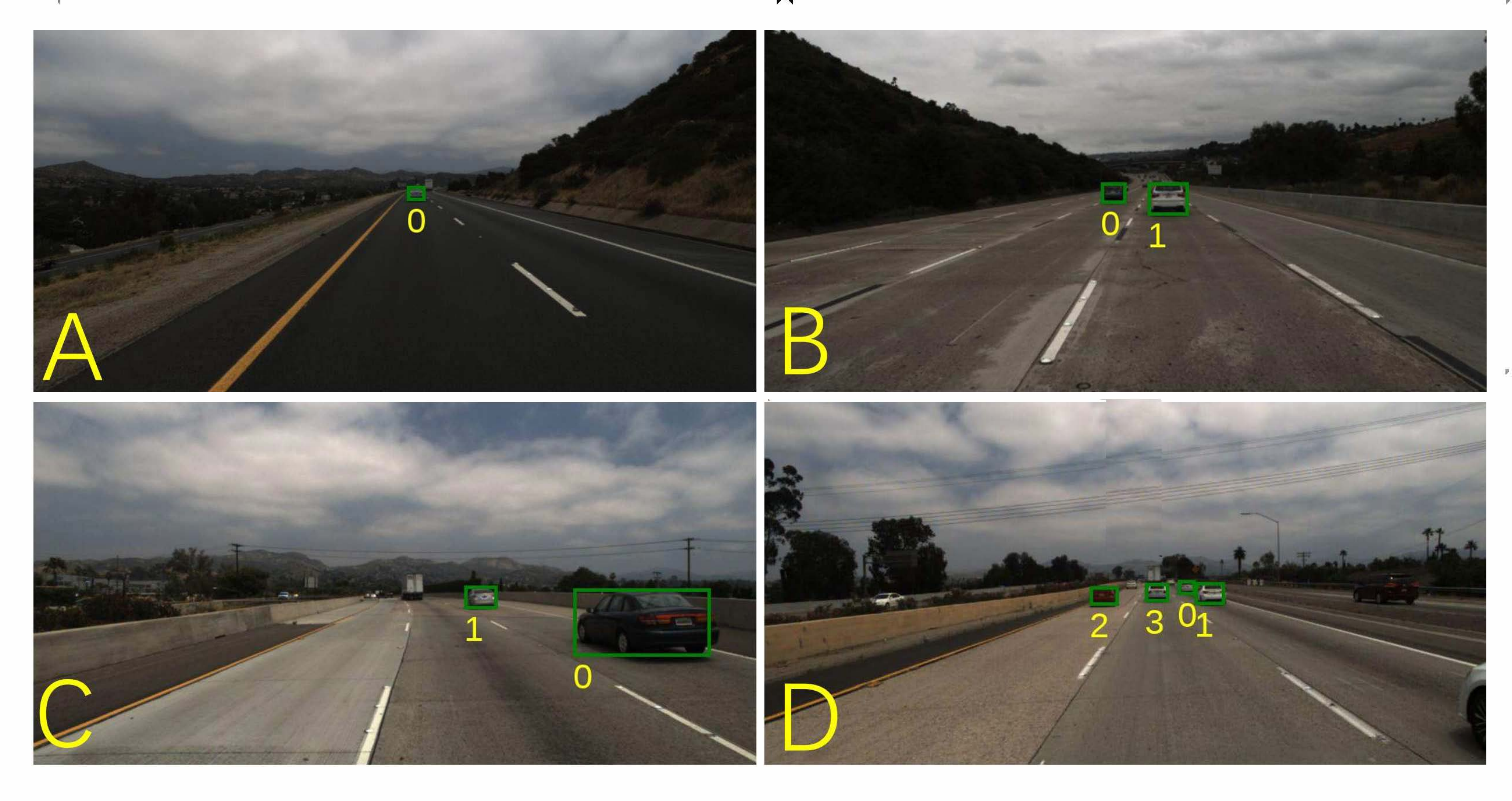}
\caption{Qualitative results on Tusimple dataset. The predicted position and velocity along (${z}$-axis, ${x}$-axis) are given in Table. \ref{tb.result} as well as corresponding ground truth.}
\label{fig.result}
\end{figure}

\begin{table}[htbp]
  \centering
  \caption{Qualitative results of \emph{\textbf{ours} full} model corresponding to Fig. \ref{fig.result}.}
  \label{tb.result}
  \begin{tabular}{c|c|c|c|c}
    \hline
    \multirow{2}*{vehicle ID} & \multicolumn{2}{c}{Position(m)} & \multicolumn{2}{|c}{Velocity(m/s)}\\
                             \cline{2-5}
		                  ~ & prediction & ground truth & prediction  & ground truth\\
    \hline
    A-0 &    (51.8, 0.6) &   (50.0, -0.2) &  (1.3, 0)  & (1.0, 0)\\
    \hline
    B-0 &    (41.2, -3.1) &    (41.0, -3.4) &  (2.8, 0)  & (2.8, 0)\\
    \hline
    B-1 &    (23.3, 0.8) &    (22.9, -0.4) &   (1.1, 0)  & (0.7, 0)\\
    \hline
    C-0 &    (9.9, 4.4) &     (9.8, 3.3) &    (-1.2, 0)  & (-0.7, 0)\\
    \hline
    C-1 &    (34.5, 4.2) &    (34.0, 2.7) &   (-0.2, 0)  & (-0.5, 0)\\
    \hline
    D-0 &    (56.8, 4.1) &     (56.3, 2.7) &  (-1.1, 0)  & (-1.2, 0)\\
    \hline
    D-1 &    (33.9, 4.0) &     (32.7, 2.5) &  (-1.3, 0)  & (-1.2, 0)\\
    \hline
    D-2 &    (34.7, -3.2) &    (33.4, -3.6) &   (4.6, 0)  & (4.5, 0)\\
    \hline
    D-3 &    (39.7, 0.5) &     (39.8, -0.3) &   (1.5, 0)  & (1.3, 0)\\
    \hline
  \end{tabular}
\end{table}

\subsection{Results on Tusimple}
 In this dataset, ground truth annotation of vehicles' position and velocity is given by two directions along ${z}$-axis and ${x}$-axis of the camera coordinate system respectively.
To evaluate our result, the output of our network is setted as a 3D vector: 1D distance and 2D velocity.
The other dimension of position is calculated by inverse perspective projection using the predicted distance and central pixel coordinate of the bounding box.
Following the Tusimple evaluation metric, we report the MSE of position and velocity in three distance range groups.
In order to compare with other state-of-art methods from the challenge, we bring the challenge leader board from \cite{kampelmuhler2018camera}.
The method introduced in \cite{kampelmuhler2018camera} won the challenge using individual models trained on all three ranges separately, while incorporating tracking, flow and depth features.
\emph{Rank2} method takes tracking bounding box as the only clue to calculate the velocity with multi-layer perceptrons and \emph{rank3} method is not released.
Unlike \emph{Rank1} using alternative models for different range groups, our model is the same for every range group, which demonstrates the robustness of our method.
\emph{\textbf{ours} org} model performs on the original image and utilizes all clues we have, including geometric clue, deep feature clue and optical flow clue;
\emph{\textbf{ours} full} is our vehicle-centric model which reduces the motion effects and perspective influences.

{\bf Performance on Benchmark  }
Since methods on this leader board did not provide the position estimation results, we focus on the comparison of velocity results and our distance regression model is analysed later on the KITTI dataset.
As shown in Table. \ref{tb.leaderboard}, \emph{\textbf{ours} org} method already reaches competitive results (average mean square error as 1.50) with the \emph{Rank2} method.
Notably, in \emph{\textbf{ours} full} model, the accuracy of velocity estimation improves significantly in all three range groups, especially for far group.
The average mean square velocity error of our approach is less than ${0.86 m^2/s^2}$ (corresponding to about ${0.48 m/s}$ absolute error), while the MSE of position is ${10.23 m^2}$. 
In addition, \emph{\textbf{ours} full} model runs in real time consuming only ${16 ms}$ for each vehicle-centric patch on a single GTX 1080ti.
Fig. \ref{fig.result} and Table. \ref{tb.result} illustrate some visual results of \emph{\textbf{ours} full} model.

\begin{figure}[htbp]
\centering
\includegraphics[width=0.5\textwidth]{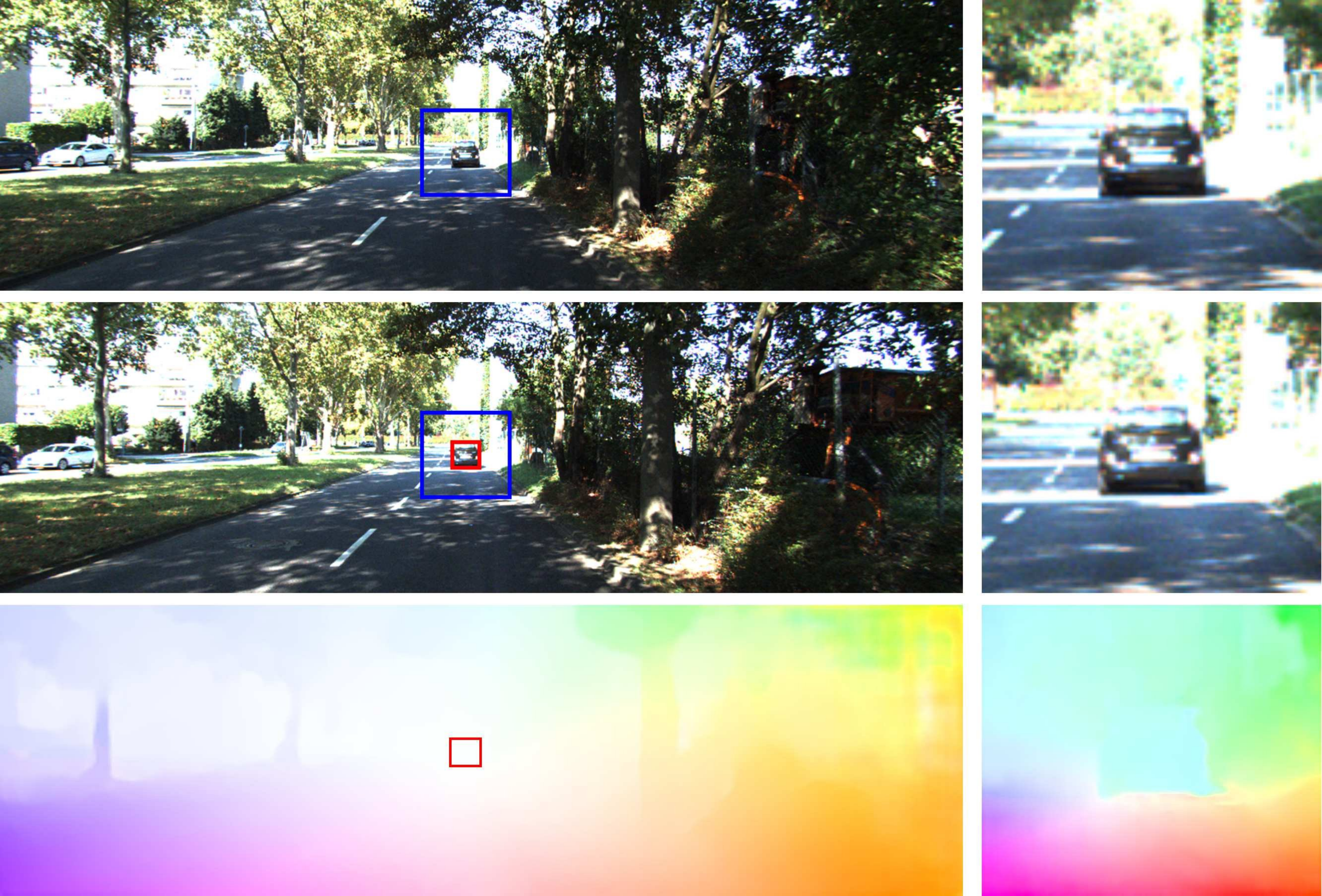}
\caption{Flow results comparison. The left column is the original images and predicted flow; The right column is the re-sampled images and predicted flow. The vehicle bounding box is labelled in red and the re-sampling area is labelled blue.}
\label{fig.flow}
\end{figure}

{\bf Vehicle-centric Sampling Impact  }
The only difference between our two methods is the vehicle-centric sampling step before processing the network.
We conduct ablation investigation into the impact of this re-sampling step on deep feature clue and flow clue.
Due to no change of the geometric clue, the distance regression model depends only on deep feature clue.
The distance regression performance therefore reflects the influence of re-sampling step on the deep feature clue.
We evaluate the distance regression accuracy in Table. \ref{tb.deepfeature}, which indicates that the accuracy changes little when inputting original images or re-sampled images.
The vehicle-centric sampling step is thought to have little influence on the deep feature clue.
By contrast, the re-sampling step helps reduce the average MSE of velocity a lot from 1.50 to 0.86 in Table. \ref{tb.leaderboard}.
Since optical flow is an essential clue for velocity estimation, the re-sampling step is considered to be beneficial for optical flow clue.
In Fig. \ref{fig.flow}, we visualize the optical flow of both original and re-sampled images in a typical scene using pre-trained PWC-Net.
One can observe that the flow of vehicle in re-sampled images is notably better than that in original images, as a vehicle moving in the same direction has small real flows.
These results clearly validate the effectiveness and importance of the vehicle-centric sampling step.
\begin{table}[htbp]
  \centering
  \caption{Vehicle distance estimation on Tusimple. \emph{\textbf{ours} org} and \emph{\textbf{ours} full} take original image and re-sampled image as input respectively.The first four metrics(error) are the lower the better; the last three metrics(accuracy) are the higher the better.}
  \label{tb.deepfeature}
  \begin{tabular}{c|c|c}
    \hline
    Metrics & \emph{\textbf{ours} org} & \emph{\textbf{ours} full} \\
    \hline
    AbsRel  & 0.037 &    0.041 \\
    \hline
    SqRel  & 0.132 &    0.152\\
    \hline
    RMS & 2.700 &     2.849 \\
    \hline
    RMSlog & 0.059 &   0.062 \\
    \hline
    ${< {1.25}^1}$ &  0.989 &   0.987  \\
    \hline
    ${< {1.25}^2}$ &  1.00 &   1.00\\
    \hline
   ${< {1.25}^3}$ &  1.00 &    1.00\\
   \hline
  \end{tabular}
\end{table}

\subsection{Results on KITTI}
As KITTI provides corresponding point clouds for each image, we can obtain the ground truth of the vehicles' distance.
The results of distance regression model are therefore emphatically analysed in this section and the velocity estimation results are simply reported in Table. \ref{tb.kittivelocity}.
There are two benchmarks related to our distance regression task on KITTI: 3D object detection and depth prediction.
We utilize methods submitted on these two benchmarks to compare with our results:
a 3D bounding box estimation algorithm(\emph{3Dbbox} \cite{mousavian20173d}) and two monocular depth regression algorithms (supervised \emph{DORN} \cite{FuCVPR18-DORN}, unsupervised \emph{Unsfm} \cite{bian2019depth}).
To evaluate \emph{3Dbbox}, the ground truth position is computed by the mean of the vehicle's point clouds referring to the center of predicted 3D bounding box.
To evaluate \emph{DORN} and \emph{Unsfm}, the minimal distance inside the vehicle bounding box is picked in both ground truth point clouds and predicted depth map.
Since depth maps from monocular depth estimation are up to an unknown scale factor, the proportion between the median of predicted distances and the median of ground truth distances is applied to provide the correct scale.

In order to analyse the efficiency of our distance regression model, we trained an independent network only based on the PWC-Net encoder without vehicle-centric operation.
Deep feature clue and two geometric supplementary clue are feed into 4 fully connect layers to estimate the distance to vehicles.
The training data consists of 3800 frames from KITTI raw dataset with at least one moving vehicle in each frame and 764 frames constitutes the testing data.
From Table.~\ref{tb.depth}, our method outperforms the \emph{3Dbbox} and \emph{Unsfm} in every metric.
Note that \emph{Unsfm} gets the worst results, because it is trained using camera ego-motion and the image warping consistency which is not satisfied for a dynamic scene with moving vehicles.
As the first place method on KITTI depth prediction benchmark, \emph{DORN} shows remarkable performances, yet our method achieves competitive results and gets less outliers than it.

\begin{table}[htbp]
  \centering
  \caption{Velocity estimation results on KITTI.}
  \label{tb.kittivelocity}
  \begin{tabular}{c|c|c|c|c}
    \hline
    ~ & MSE(near) & MSE(medium)  & MSE(far) & MSE(average)\\
    \hline
    \emph{\textbf{ours} full} &  0.29 &    0.93 &    1.57 &   0.94 \\
    \hline
    \end{tabular}
\end{table}

\begin{table}[htbp]
  \centering
  \caption{Distance estimation on KITTI. \emph{3Dbbox}\cite{mousavian20173d} is a 3D object detection network; \emph{DORN}\cite{FuCVPR18-DORN} and \emph{Unsfm}\cite{bian2019depth} are depth prediction networks; \emph{\textbf{ours}} is our distance regression network.}
  \label{tb.depth}
  \begin{tabular}{c|c|c|c|c}
    \hline
    Metrics & \emph{3Dbbox}\cite{mousavian20173d} & \emph{DORN}\cite{FuCVPR18-DORN} & \emph{Unsfm}\cite{bian2019depth} & \emph{\textbf{ours}}\\
    \hline
    AbsRel  & 0.222 &    0.078 &     0.219 &    \textbf{0.075}\\
    \hline
    SqRel  & 1.863 &    0.505&      1.924 &    \textbf{0.474}\\
    \hline
    RMS & 7.696 &     \textbf{4.078}&     7.873 &     4.639 \\
    \hline
    RMSlog & 0.228 &   0.179&    0.338&    \textbf{0.124} \\
    \hline
    ${< {1.25}^1}$ &  0.659 &    \textbf{0.927} & 0.710&      0.912 \\
    \hline
    ${< {1.25}^2}$ &  0.966 &    0.985&     0.886&  \textbf{0.996} \\
    \hline
   ${< {1.25}^3}$ &  0.994 &    0.995&    0.933 &  \textbf{1.000} \\
    
   \hline
    \end{tabular}
\end{table}

\section{Conclusion}
In this paper, we have presented an end-to-end trainable deep neural network for inter-vehicle distance and relative velocity estimation in ADAS applications.
The network is based on a light-weight pipeline which focuses on the extraction of both spatial and temporal clue from two images of a monocular sequence, without the need of vehicle tracking or road segmentation.
Tested on standard benchmark datasets, our method outperforms existing state of the art methods. The obtained estimations are adequate for forward collision warning etc. applications.
One possible future research direction is to explore the 3D model fitting and ego-motion estimation of other vehicles in video, to facilitate precise 3D pose and motion estimation of the lead vehicles.

\bibliographystyle{IEEEtran}

\bibliography{survey}

\end{document}